# Symbolic Regression Methods for Reinforcement Learning

**JIŘÍ KUBALÍK[1], ERIK DERNER[1,2], (Student Member, IEEE), JAN ŽEGKLITZ[2], AND ROBERT BABUŠKA[1,3], (Member, IEEE)**

[1]Czech Institute of Informatics, Robotics, and Cybernetics, Czech Technical University in Prague, 16000 Prague, Czech Republic
[2]Faculty of Electrical Engineering, Czech Technical University in Prague, 16627 Prague, Czech Republic
[3]Cognitive Robotics, Delft University of Technology, 2628 Delft, The Netherlands

Corresponding author: Jiří Kubalík (jiri.kubalik@cvut.cz)

This work was supported in part by the European Regional Development Fund under the project Robotics for Industry 4.0 under Grant CZ.02.1.01/0.0/0.0/15_003/0000470, in part by the Grant Agency of the Czech Republic (GAČR) titled ‘‘Symbolic Regression for Reinforcement Learning in Continuous Spaces’’ under Grant 15-22731S, and in part by the Grant Agency of the Czech Technical University in Prague under Grant SGS19/174/OHK3/3T/13.

**ABSTRACT** Reinforcement learning algorithms can solve dynamic decision-making and optimal control problems. With continuous-valued state and input variables, reinforcement learning algorithms must rely on function approximators to represent the value function and policy mappings. Commonly used numerical approximators, such as neural networks or basis function expansions, have two main drawbacks: they are black-box models offering little insight into the mappings learned, and they require extensive trial and error tuning of their hyper-parameters. In this paper, we propose a new approach to constructing smooth value functions in the form of analytic expressions by using symbolic regression. We introduce three off-line methods for finding value functions based on a state-transition model: symbolic value iteration, symbolic policy iteration, and a direct solution of the Bellman equation. The methods are illustrated on four nonlinear control problems: velocity control under friction, one-link and two-link pendulum swing-up, and magnetic manipulation. The results show that the value functions yield well-performing policies and are compact, mathematically tractable, and easy to plug into other algorithms. This makes them potentially suitable for further analysis of the closed-loop system. A comparison with an alternative approach using neural networks shows that our method outperforms the neural network-based one.

## I. INTRODUCTION

Reinforcement learning (RL) in continuous-valued state and input spaces relies on function approximators. Various types of numerical approximators have been used to represent the value function and policy mappings: expansions with fixed or adaptive basis functions [1], [2], regression trees [3], local linear regression [4], [5], and deep neural networks [6]–[10].

The choice of a suitable approximator, in terms of its structure and hyperparameters (number, type, and distribution of the basis functions, number and size of layers in a neural network, etc.), is an ad hoc step that requires extensive trial and error tuning. There are no guidelines for designing a good value-function approximator. A large amount of expert knowledge and haphazard tuning is required when applying RL techniques to continuous-valued problems. In addition, these approximators are black-box models, yielding limited insight and possibilities for analysis. Moreover, approaches based on deep neural networks often suffer from the lack of reproducibility [11]. Finally, the interpolation properties of numerical function approximators may adversely affect the control performance and result in chattering control signals and steady-state errors [12]. In practice, this often makes RL inferior to alternative control design methods, despite the theoretical potential of RL to produce optimal control policies [13].

To overcome these limitations, we propose a novel approach that uses symbolic regression (SR) to construct an analytic representation of the value function automatically. SR has been used in nonlinear data-driven modeling with quite impressive results [14]–[17]. To our best knowledge, there have been no reports in the literature on the use of SR for constructing value functions. The closest related research employs genetic programming for fitting already available value functions [18], [19]. The authors of [18] use GP to find an algebraic expression that fits the sample points of a value function previously obtained via value iteration. The result in [19] relies on the fact that the so-called threshold policy for the MDP is known a priori, and GP produces a description of this threshold policy in terms of the MDP parameters. In both cases, the task is to fit an algebraic expression to a set of data sampled from some known value or policy function. This approach is different from our case, where the task is to construct the V-function in the form of an analytic expression based on the raw data sampled by using a state-transition model of the system to be controlled. Recently, several other works on generating RL policies through genetic programming have been published. In [20], the authors propose a Cartesian Genetic Programming (CGP) to evolve control policies for Atari games using raw pixel input and discrete actions. To efficiently cope with the input format, the CGP uses a function set designed specifically for computer vision – with list processing, matrix, and vector operations. Standard tree-based GP using basic arithmetic and logical functions and operators was proposed in [21] and [22] to evolve policies in the form of algebraic equations for RL problems with continuous state and action spaces. Experiments on three reinforcement learning benchmarks show that the method can learn well-performing policies that are of low complexity compared to the neural network-based ones.

The paper is organized as follows. Section II describes the reinforcement learning framework considered in this work. Section III presents the proposed symbolic methods: symbolic value iteration, symbolic policy iteration, and a direct solution of the Bellman equation. Section IV presents the experimental results obtained with the proposed methods on four nonlinear control problems: velocity control under nonlinear friction, one-link and two-link pendulum swing-up, and magnetic manipulation. Performance of the methods and other related aspects are discussed in Section V. Finally, Section VI concludes the paper.

## II. RL FRAMEWORK

The dynamic system of interest is described by the state transition function

$$x_{k+1} = f(x_k, u_k) \tag{1}$$

with $x_k, x_{k+1} \in \mathcal{X} \subset \mathbb{R}^n$ and $u_k \in \mathcal{U} \subset \mathbb{R}^m$, where subscript $k$ denotes discrete time instants. Function $f$ is assumed to be given, but it does not have to be stated by explicit equations; it can be, for instance, a generative model given by a numerical simulation of complex differential equations. The control goal is specified through a *reward function* which assigns a scalar reward $r_{k+1} \in \mathbb{R}$ to each state transition from $x_k$ to $x_{k+1}$:

$$r_{k+1} = \rho(x_k, u_k, x_{k+1}) \,. \tag{2}$$

This function is defined by the user and typically calculates the reward based on the distance of the current state to a given reference (goal) state $x_r$. The state transition model and the associated reward function form the Markov decision process (MDP).

The goal of RL is to find an optimal control policy $\pi : \mathcal{X} \to \mathcal{U}$ that for each initial state $x_0$ selects a control action so that the cumulative discounted reward over time, called the return, is maximized:

$$R^\pi(x_0) = \sum_{k=0}^{\infty} \gamma^k \rho\big(x_k, \pi(x_k), x_{k+1}\big) \,. \tag{3}$$

Here $\gamma \in (0, 1)$ is the discount factor. The return is approximated by the value function $V^\pi : X \to \mathbb{R}$ defined as:

$$V^\pi(x) = \sum_{k=0}^{\infty} \gamma^k \rho\big(x_k, \pi(x_k), x_{k+1}\big), \;\; x_0 = x \,. \tag{4}$$

An approximation of the optimal V-function, denoted by $\hat{V}^*(x)$, can be computed by solving the Bellman optimality equation

$$\hat{V}^*(x) = \max_{u\in\mathcal{U}} \Big[\rho\big(x, u, f(x, u)\big) + \gamma \hat{V}^*\big(f(x, u)\big)\Big] \,. \tag{5}$$

To simplify the notation, we drop the hat and the star superscript: $V(x)$ will therefore denote the approximately optimal V-function. Based on $V(x)$, the optimal control action in any given state $x$ is found as the one that maximizes the right-hand side of (5):

$$\pi(x) = \operatorname*{argmax}_{u\in\mathcal{U}} \, \big[\rho\big(x, u, f(x, u)\big) + \gamma V\big(f(x, u)\big)\big] \,. \tag{6}$$

In this paper, we use the above RL framework based on V-functions. However, the proposed methods can be applied to Q-functions as well.[1]

## III. SOLVING BELLMAN EQUATION BY SYMBOLIC REGRESSION

We employ SR to construct an analytic approximation of the value function. Symbolic regression is based on genetic programming, and its purpose is to find an analytic equation describing given data. Our specific objective is to find an analytic equation for the value function that satisfies the Bellman optimality equation (5). SR appears to be a suitable technique for this task: it does not rely on any prior knowledge on the form of the value function, which is generally unknown. It also has the potential to provide more compact representations than, for instance, deep neural networks or basis function expansion models. In this work, we employ two different SR methods: a variant of Single Node Genetic Programming [23]–[26] and a variant of Multi-Gene Genetic Programming [27]–[29]. In the sequel, we use the term 'symbolic V-function' to denote the V-function model constructed by SR.

[1]The Q-function $Q^\pi(x, u)$ approximates the return obtained when first taking action $u$ in state $x$ and then following policy $\pi$ until the end of the episode.

### A. SYMBOLIC REGRESSION

SR methods were reported to perform best when using a linear combination of nonlinear functions constructed by means of genetic algorithms [30], [31]. Following this approach, we define the class of symbolic V-functions as:

$$V(x) = \beta_0 + \sum_{i=1}^{n_f} \beta_i \varphi_i(x). \tag{7}$$

The nonlinear functions $\varphi_i(x)$, called features, are constructed through genetic programming using a predefined set of elementary functions $\mathcal{F}$ specified by the user. These functions can be nested, and the SR algorithm evolves their combinations by using standard evolutionary operations such as mutation. The complexity of the symbolic V-function is constrained by two user-defined parameters: the maximum number of features in the symbolic model and the maximum feature tree depth. Coefficients $\beta$ are estimated by least squares, with or without regularization.

### B. DATA SET

To apply SR, we first generate a set of $n_x$ discrete states sampled from $\mathcal{X}$:

$$X = \{x_1, \ldots, x_{n_x}\} \subset \mathcal{X},$$

and a set of $n_u$ discrete control inputs sampled from $\mathcal{U}$:

$$U = \{u_1, \ldots, u_{n_u}\} \subset \mathcal{U}\,.$$

The generic training data set for SR is then given by:

$$D = \{d_1, \ldots, d_{n_x}\} \tag{8}$$

where each training sample $d_i$ is the tuple:

$$d_i = \langle x_i, x_{i,1}, r_{i,1}, \ldots, x_{i,n_u}, r_{i,n_u} \rangle$$

consisting of the state $x_i \in X$, all the next states $x_{i,j}$ obtained by applying in $x_i$ all the control inputs $u_j \in U$ to the system model (1), and all the corresponding rewards $r_{i,j} = \rho\big(x_i, u_j, f(x_i, u_j)\big)$.

In the sequel, $V$ denotes the analytic representation of the value function generated by SR applied to the data set $D$ and we present three different approaches to solving the Bellman equation.

### C. DIRECT SYMBOLIC SOLUTION OF BELLMAN EQUATION

This approach (`direct`) evolves the symbolic value function so that it satisfies (5). The optimization criterion (fitness function) is the mean-squared error between the left-hand side and right-hand side of the Bellman equation, i.e., the Bellman error (residual) over all the training samples in $D$:

$$J^{\text{direct}} = \frac{1}{n_x} \sum_{i=1}^{n_x} \Big[ \max_j \big( r_{i,j} + \gamma \underbrace{V(x_{i,j})}_{\text{evolved}} \big) - \underbrace{V(x_i)}_{\text{evolved}} \Big]^2 . \tag{9}$$

Unfortunately, the problem formulated in this way proved too hard to be solved by SR, as illustrated later in Section IV. We hypothesize that this difficulty stems from the fact that the fitness of the value function to be evolved[2] is evaluated through the complex implicit relation in (9), which is not a standard regression problem. In other words, no target is given to which the value function should be fitted. By modifying SR, the problem might be rendered feasible. Still, in this paper, we successfully adopt an iterative approach, leading to the symbolic value iteration and symbolic policy iteration, as described below.

### D. SYMBOLIC VALUE ITERATION

In symbolic value iteration (`SVI`), the optimal value function is found iteratively, just like in standard value iteration [32]. In each iteration $\ell$, the value function $V_{\ell-1}$ from the previous iteration is used to compute the target for improving the value function $V_\ell$ in the current iteration. For each state $x_i \in X$, the target $t_{i,\ell} \in \mathbb{R}$ is calculated by evaluating the right-hand-side of (5):

$$t_{i,\ell} = \max_{u \in U} \Big( \rho(x_i, u, f(x_i, u)) + \gamma V_{\ell-1}\big(f(x_i, u)\big) \Big). \tag{10}$$

Here, the maximization is carried out over the predefined discrete control action set $U$. Note that virtually all control systems use discrete control actions – either due to digital-to-analog conversion or due to the nature of the actuator itself, e.g., a stepping motor. As the sensitivity of the control loop to discrete control actions is low (approximately the reciprocal of the loop gain), most control loops tolerate even a small number of control actions, as long as the action corresponding to the desired goal state is included in the control action set. In principle, it would also be possible to use numerical or even symbolic optimization over the original continuous set $\mathcal{U}$. However, this is computationally more expensive, as the optimization problem would have to be solved $n_x$ times at the beginning of each iteration. For this reason, we prefer the maximization over $U$, as stated in (10). In addition, as the next states and rewards are pre-computed and provided to the `SVI` algorithm in the data set $D$ (8), we can replace (10) by its computationally more efficient equivalent:

$$t_{i,\ell} = \max_j \big( r_{i,j} + \gamma V_{\ell-1}(x_{i,j}) \big). \tag{11}$$

Given the target $t_{i,\ell}$, an improved value function $V_\ell$ is constructed by applying SR with the following fitness function:

$$J_\ell^{\text{SVI}} = \frac{1}{n_x} \sum_{i=1}^{n_x} \Big[ \underbrace{t_{i,\ell}}_{\text{target}} - \underbrace{V_\ell(x_i)}_{\text{evolved}} \Big]^2 . \tag{12}$$

This fitness function is again the mean-squared Bellman error. However, as opposed to (9), the above criterion (12) defines a true regression problem: the target to be fitted is fixed as it is based on $V_{\ell-1}$ from the previous iteration. In the first iteration, $V_0$ can be initialized either by some suitable function or as $V_0(x) = 0$ for all $x \in \mathcal{X}$, in the absence of a better initial value. In the latter case, the first target becomes the largest reward over all the next states.

[2]The term evolved refers to the fact that the approximator of the given function is constructed through SR, i.e., evolved using genetic programming.

In each iteration, the training data set for SR is composed as follows:

$$D_\ell^{\text{SVI}} = \{d_1, \dots, d_{n_x}\} \text{ with } d_i = \langle x_i, t_{i,\ell} \rangle$$

i.e., each sample contains the state $x_i$, and the corresponding target $t_{i,\ell}$ computed by (11).

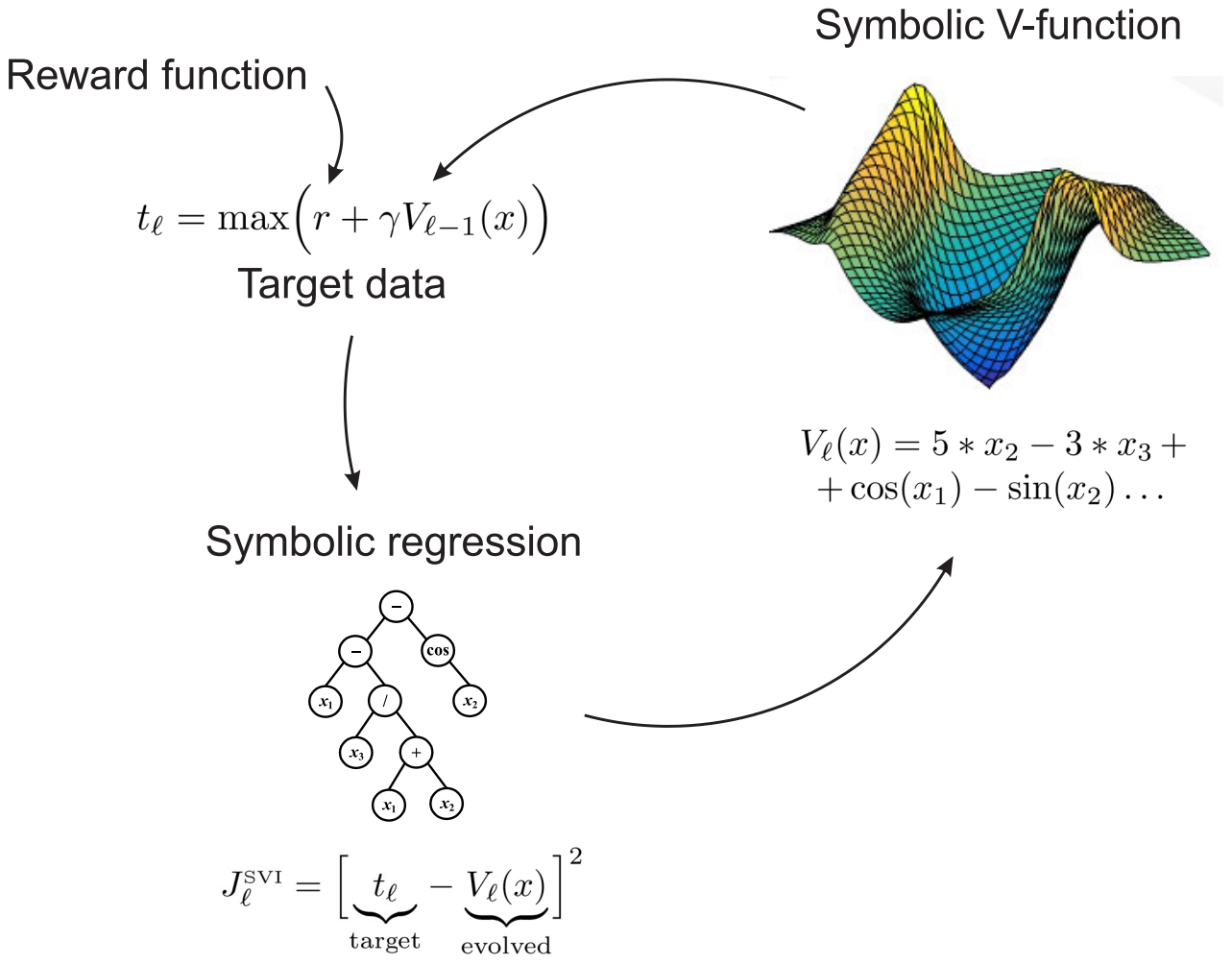

**FIGURE 1. Symbolic value iteration loop. In each iteration, the target data for SR are computed using the Bellman equation right-hand side. SR then improves the value function, and the process repeats.**

The SVI procedure terminates once a predefined maximum number of iterations $n_i$ has been reached. Other stopping criteria can be employed, such as terminating the iteration when the following condition is satisfied:

$$\max_i |V_\ell(x_i) - V_{\ell-1}(x_i)| \le \epsilon \tag{13}$$

with $\epsilon$ a user-defined convergence threshold. The resulting *symbolic value iteration* algorithm is given in Algorithm 1 and depicted in Figure 1. In each iteration, the SR algorithm is run for $n_g$ generations.

**Algorithm 1** Symbolic Value Iteration (SVI)

**Input**: training data set $D$, $n_i$
$\ell \leftarrow 0, \;\; V_0(x) = 0, \;\; \forall x \in \mathcal{X}$
**while** $\ell < n_i$ **do**
  $\ell \leftarrow \ell + 1$
  $\forall x_i \in X$ compute $t_{i,\ell}$ by using (11)
  $D_\ell^{\text{SVI}} \leftarrow \{d_1, \dots, d_{n_x}\}$ with $d_i = \langle x_i, t_{i,\ell} \rangle$
  $V_\ell \leftarrow \text{SymbolicRegression}(D_\ell^{\text{SVI}}, J_\ell^{\text{SVI}})$
**end**
$V \leftarrow V_\ell$
**Output**: Symbolic value function $V$

### E. SYMBOLIC POLICY ITERATION

The symbolic policy iteration (SPI) algorithm iteratively improves the V-function estimate. However, rather than using $V_{\ell-1}$ to compute the target in each iteration, we derive from $V_{\ell-1}$ the currently optimal policy and plug it into the Bellman equation, so eliminating the maximum operator.

Given the value function $V_{\ell-1}$ from the previous iteration, for each state $x_i \in X$, the corresponding currently optimal control action $u_i^*$ is computed by:

$$u_i^* = \operatorname*{argmax}_{u \in U} \Big(\rho(x_i, u, f(x_i, u)) + \gamma V_{\ell-1}\big(f(x_i, u)\big)\Big), \tag{14}$$

$\forall x_i \in X$. Again, the maximization could be carried out over the original continuous set $\mathcal{U}$, rather than the discrete set $U$, which would incur higher computational costs.

Now, for each state $x_i$ and the corresponding optimal control action $u_i^*$, the optimal next state $x_i^*$ and the respective reward $r_i^*$ can be computed:

$$x_i^* = f(x_i, u_i^*), \quad r_i^* = \rho(x_i, u_i^*, x_i^*)\,. \tag{15}$$

As the next states and rewards are provided to the SPI algorithm in the data set $D$ (8), we can replace (14) by its computationally more efficient equivalent. The index $j^*$ of the optimal control action selected from $U$ is found by

$$j^* = \operatorname*{argmax}_j \big(r_{i,j} + \gamma V_{\ell-1}(x_{i,j})\big)\,, \tag{16}$$

$$x_i^* = x_{i,j^*}, \quad r_i^* = r_{i,j^*} \tag{17}$$

with $x_{i,j^*}$ and $r_{i,j^*}$ selected from $D$. Given these samples, we can now construct the training data set for SR as follows:

$$D_\ell^{\text{SPI}} = \{d_1, \dots, d_{n_x}\} \text{ with } d_i = \langle x_i, x_i^*, r_i^* \rangle\,.$$

This means that each sample $d_i$ contains the state $x_i$, the currently optimal next state $x_i^*$ and the respective reward $r_i^*$. In each iteration $\ell$ of SPI, an improved approximation $V_\ell$ is sought by means of SR with the following fitness function:

$$J_\ell^{\text{SPI}} = \frac{1}{n_x}\sum_{i=1}^{n_x}\Big(\underbrace{r_i^*}_{\text{target}} - [\underbrace{V_\ell(x_i)}_{\text{evolved}} - \gamma \underbrace{V_\ell(x_i^*)}_{\text{evolved}}]\Big)^2\,. \tag{18}$$

The fitness is again the mean-squared Bellman error, where only the currently optimal reward serves as the target for the difference $V_\ell(x_i) - \gamma V_\ell(x_i^*)$, with $V_\ell$ evolved by SR. The resulting *symbolic policy iteration* algorithm is given in Algorithm 2.

### F. PERFORMANCE MEASURES FOR EVALUATING VALUE FUNCTIONS

Note that the convergence of the iterative algorithms is not necessarily monotonic w.r.t. the mean-squared Bellman error, see Figure 3b. This behavior is similar to other approximation-based methods like the fitted Q-iteration algorithm [3]. Therefore, it is not meaningful to retain only the last solution. Instead, we store the intermediate solutions from all iterations and use a posteriori analysis to select the best value function according to the performance measures described below.

**Algorithm 2** Symbolic Policy Iteration (SPI)

**Input**: training data set $D$, $n_i$
$\ell \leftarrow 0,\ V_0(x) = 0,\ \ \forall x \in \mathcal{X}$
**while** $\ell < n_i$ **do**
  $\ell \leftarrow \ell + 1$
  $\forall x_i \in X$ select $x_i^*$ and $r_i^*$ from $D$ by (16) and (17)
  $D_\ell^{\text{SPI}} \leftarrow \{d_1, \ldots, d_{n_x}\}$ with $d_i = \langle x_i, x_i^*, r_i^* \rangle$
  $V_\ell \leftarrow \text{SymbolicRegression}(D_\ell^{\text{SPI}}, J_\ell^{\text{SPI}})$
**end**
$V \leftarrow V_\ell$
**Output**: Symbolic value function $V$

**Root mean squared Bellman error** (BE) is calculated over all $n_x$ state samples in the training data set $D$ according to

$$\text{BE} = \sqrt{\frac{1}{n_x}\sum_{i=1}^{n_x}\Big[\max_j\big(r_{i,j} + \gamma V(x_{i,j})\big) - V(x_i)\Big]^2}.$$

In the optimal case, the Bellman error is equal to zero.

The following two measures are calculated based on closed-loop control simulations with the state transition model (1). The simulations start from $n_s$ different initial states in the set $X_{\text{init}}$ ($n_s = |X_{\text{init}}|$) and run for a fixed amount of time $T_{\text{sim}}$. At each simulation time step, the optimal control action is computed according to the argmax policy (6).

**Mean discounted return** ($\text{R}_\gamma$) is calculated over the simulations from all the initial states in $X_{\text{init}}$:

$$R_\gamma = \frac{1}{n_s}\sum_{s=1}^{n_s}\sum_{k=0}^{T_{\text{sim}}/T_s} \gamma^k \rho\big(x_k^{(s)}, \pi(x_k^{(s)}), x_{k+1}^{(s)}\big)$$

where $(s)$ denotes the index of the simulation initialized at $x_0^{(s)} \in X_{\text{init}}$ and $T_s$ is the sampling period. Larger values of $\text{R}_\gamma$ indicate better performance.

**Percentage of successful simulations** ($S$) within all $n_s$ simulations is calculated as

$$S = 100\frac{n_{succ}}{n_s}\ \%\,,$$

where $n_{succ}$ is the number of successful simulations. A simulation is considered successful if the state $x$ reaches a predefined neighborhood of the goal state and stays there for the final $T_{\text{end}}$ seconds of the simulation run. The goal state neighborhood $N(x_r)$ is defined using the neighborhood size parameter $\varepsilon \in \mathbb{R}^n$ as follows:

$$N(x_r) = \big\{x : |x_{r,i} - x_i| \le \varepsilon_i, \forall i \in \{1, 2, \ldots, n\}\big\}.$$

Larger values of $S$ correspond to better performance.

### G. EXPERIMENTAL EVALUATION

Each of the three proposed approaches (direct, SVI, and SPI) was implemented in two variants, one using Single Node Genetic Programming (SNGP) and the other one using Multi-Gene Genetic Programming (MGGP). Both SR methods use a linear transformation of the original input variables. In SNGP, the weights assigned to the variables are tuned purely by means of mutation operators, while in MGGP, the weights are tuned using a gradient method based on the back-propagation algorithm. A detailed explanation of the SR algorithms and their parameters is beyond the scope of this paper. We refer the interested reader for more details on the implementation of SNGP to [26] and for MGGP to [29]. A specific feature of the SNGP implementation is that it adds to the raw fitness function (i.e., (9), (12), and (18)) a penalty for a wrong position of the maximum of the V-function model. In particular, the raw fitness value is multiplied by a penalty factor $(1 + d)$, where $d$ grows linearly with the distance between the actual position of the V-function's maximum and the desired position of the maximum, which is at the goal state of the given problem.

There are six algorithms in total to be tested: direct-SNGP, direct-MGGP, SPI-SNGP, SPI-MGGP, SVI-SNGP and SVI-MGGP. Note, however, that our goal is not to compare the two symbolic regression algorithms. Instead, we want to demonstrate that the proposed symbolic RL methods are general and can be implemented by using more than one specific SR algorithm.

Each algorithm was run $n_r = 30$ times with the same parameters but with a different randomization seed. Each run delivers three winning V-functions, which are the best ones with respect to $\text{R}_\gamma$, BE and $S$, respectively. Statistics such as the median, minimum, and maximum calculated over the set of $n_r$ respective winner V-functions are used as performance measures of the particular method (SVI, SPI and direct) and the SR algorithm (SNGP, MGGP). For instance, the median of $S$ is calculated as

$$\operatorname*{med}_{r=1..n_r}\big(\max_{i=1..n_i}(S_{r,i})\big) \tag{19}$$

where $S_{r,i}$ denotes the percentage of successful simulations in iteration $i$ of run $r$. For the direct method, the above maximum is calculated over all generations of the SR run.

For comparison purposes, we have calculated a baseline solution, which is a numerical V-function approximation calculated by the fuzzy V-iteration algorithm [13] with triangular basis functions.

## IV. EXPERIMENTS

This section reports experiments for four nonlinear control problems: friction compensation, 1-DOF and 2-DOF pendulum swing-up, and magnetic manipulation. We also report experiments with neural network-based approximators (NN approximators) used instead of the symbolic V-function in the SVI method.

While the chosen test problems have low-dimensional input and state spaces, they represent challenging control problems as none of them can be solved by linear control methods. Moreover, they are challenging even for neural-network-based reinforcement learning methods, as shown in our experiments and literature; see, for example [33].

The parameters of the experiments are listed in Table 8. The SR methods worked with the following setting. The number of iterations, $n_i$, was 50 and 30 for `SVI` and `SPI`, respectively. It was smaller for `SPI` as this method converges faster. The `direct` method ran for 50 000 generations (the method does not iterate in the sense that the `SPI` and `SVI` methods do). The choice of the elementary functions used by SR to compose the nonlinear features of the models (7) can significantly affect the performance of the methods. However, optimizing the elementary function set is not the main objective of this work. Therefore, we used a rather small, yet sufficient set of elementary functions consisting of three basic algebraic operations and three univariate nonlinear functions, $\mathcal{F} = \{*, +, -, \text{square}, \text{cube}, \text{bent identity}^3\}$. This set was used by both SR methods on all test problems. The maximum number of features was set to 10, and the maximum depth of feature trees to 7. The remaining parameters of the experiment are listed in Table 8.

### A. FRICTION COMPENSATION

We start by illustrating the working of the proposed methods on a practically relevant first-order nonlinear motion-control problem. Many applications require high-precision position and velocity control, which is often hampered by the presence of friction. Without proper nonlinear compensation, friction causes significant tracking errors, stick-slip motion, and limit cycles. To address these problems, we design a nonlinear velocity controller for a DC motor with friction by using the proposed symbolic methods.

The continuous-time system dynamics are given by:

$$I\dot{v}(t) + (b + \frac{K^2}{R})v(t) + F_c\big(v(t), u(t), c\big) = \frac{K}{R}u(t) \quad (20)$$

with $v(t)$ and $\dot{v}(t)$ the angular velocity and acceleration, respectively. The angular velocity varies in the interval $[-10, 10]\,\text{rad}\cdot\text{s}^{-1}$. The control input $u \in [-4, 4]$ V is the voltage applied to the DC motor and the parameters of the system are: moment of inertia $I = 1.8 \times 10^{-4}\,\text{kg}\cdot\text{m}^2$, viscous friction coefficient $b = 1.9 \times 10^{-5}\,\text{N}\cdot\text{m}\cdot\text{s}\cdot\text{rad}^{-1}$, motor constant $K = 0.0536\,\text{N}\cdot\text{m}\cdot\text{A}^{-1}$, armature resistance $R = 9.5\,\Omega$, and Coulomb friction coefficient $c = 8.5 \times 10^{-3}\,\text{N}\cdot\text{m}$.

The Coulomb friction force $F_c$ is modeled as [34]:

$$F_c\big(v(t), u(t), c\big) = \begin{cases} c & \text{if } v(t) > 0 \text{ or } v(t) = 0 \text{ and } u(t) > c\dfrac{R}{K} \\ -c & \text{if } v(t) < 0 \text{ or } v(t) = 0 \text{ and } u(t) < -c\dfrac{R}{K} \\ \dfrac{K}{R}u(t) & \text{if } v(t) = 0 \text{ and } \left|\dfrac{K}{R}u(t)\right| \leq c \end{cases}$$

The discrete-time transitions are obtained by numerically integrating the continuous-time dynamics using the fourth-order Runge-Kutta method and a sampling period $T_s = 0.001$ s. The state is the velocity, $x = v$, and the reward function is defined as:

$$r_{k+1} = \rho(x_k, u_k, x_{k+1}) = -\sqrt{|x_r - x_k|} \quad (21)$$

with $x_r = 7\,\text{rad}\cdot\text{s}^{-1}$ the desired velocity (goal state).

In each of the 30 runs, we selected the best V-function with respect to $S$. Figure 2 shows the median values of $S$ calculated for the V-functions over all 30 runs according to (19). The `SVI` method is consistently the best one, followed by `SPI` and `direct`.

**TABLE 1. Performance of the symbolic methods on the friction compensation problem. The performance of the baseline V-function is $R_\gamma = -42.158$, BE = $1.7 \times 10^{-5}$, $S$ = 100%.**

| SNGP | `direct` | `SPI` | `SVI` |
|---|---|---|---|
| $R_\gamma$ [–] | −48.184 | −42.563 | −42.339 |
| BE [–] | 0.301 | 0.0212 | 2.571 |
| $S$ [%] | 0 | 100 | 100 |
| | [0, 100 (11)] | [0, 100 (22)] | [100, 100 (30)] |
| MGGP | `direct` | `SPI` | `SVI` |
| $R_\gamma$ [–] | −48.184 | −42.565 | −42.274 |
| BE [–] | 0.719 | 1.552 | 2.619 |
| $S$ [%] | 0 | 100 | 100 |
| | [0, 0 (30)] | [0, 100 (28)] | [100, 100 (30)] |

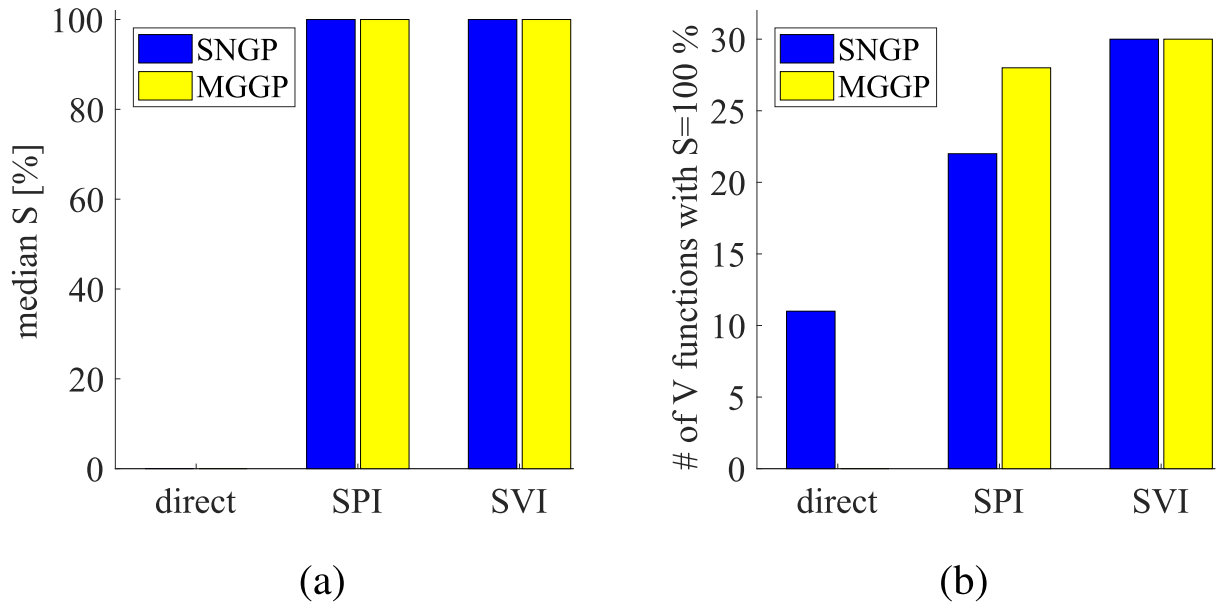

**FIGURE 2. Performance on the friction compensation problem: (a) median percentage of successful simulations $S$, (b) the number of runs in which a V-function with S = 100% was found.**

The performance measures $R_\gamma$, BE and $S$ are listed in Table 1. For the $S$ measure, the first two numbers in the square brackets are the minimum and maximum value, and the number in parentheses is the frequency of the maximum value. Interestingly, we found that low BE does not necessarily correlate with a high performance of the V-function in the control task.

Figure 3 shows examples of well-performing symbolic V-functions found through SR, compared to a baseline V-function calculated using the numerical approximator [13]. A closed-loop simulation is presented in Figure 4. Both the symbolic and baseline V-function yield optimal performance. The proposed symbolic methods reliably find well-performing V-functions for the friction compensation problem. Interestingly, even the `direct` approach can solve this problem when using the SNGP algorithm. However, it finds a well-performing V-function with respect to $S$ only in approximately one third of the runs.

[3] https://en.wikipedia.org/wiki/Bent_function

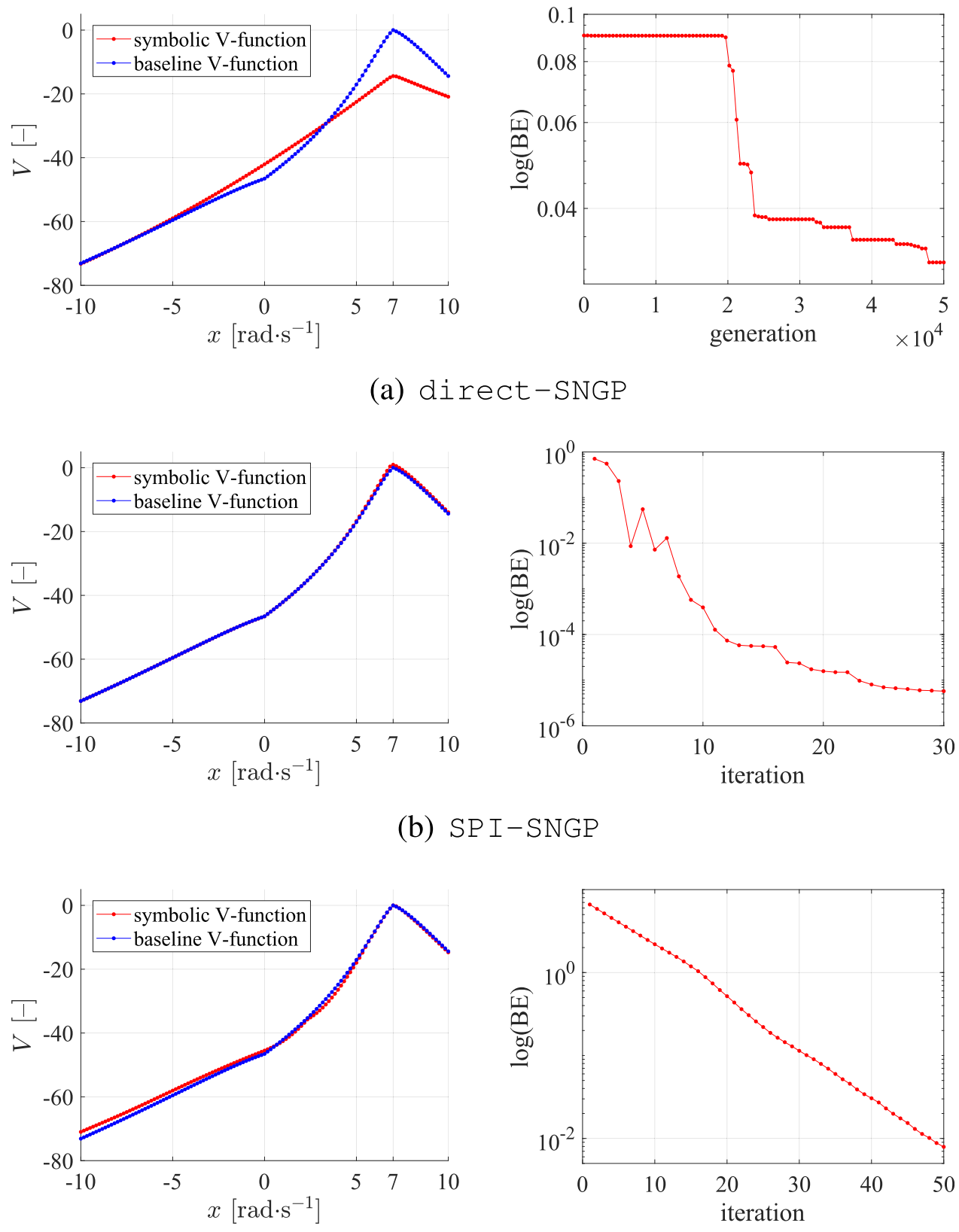

(a) direct-SNGP

(b) SPI-SNGP

(c) SVI-SNGP

**FIGURE 3. Examples of typical well-performing V-functions found for the friction compensation problem. Left: the symbolic V-function compared to the baseline. Right: the Bellman error.**

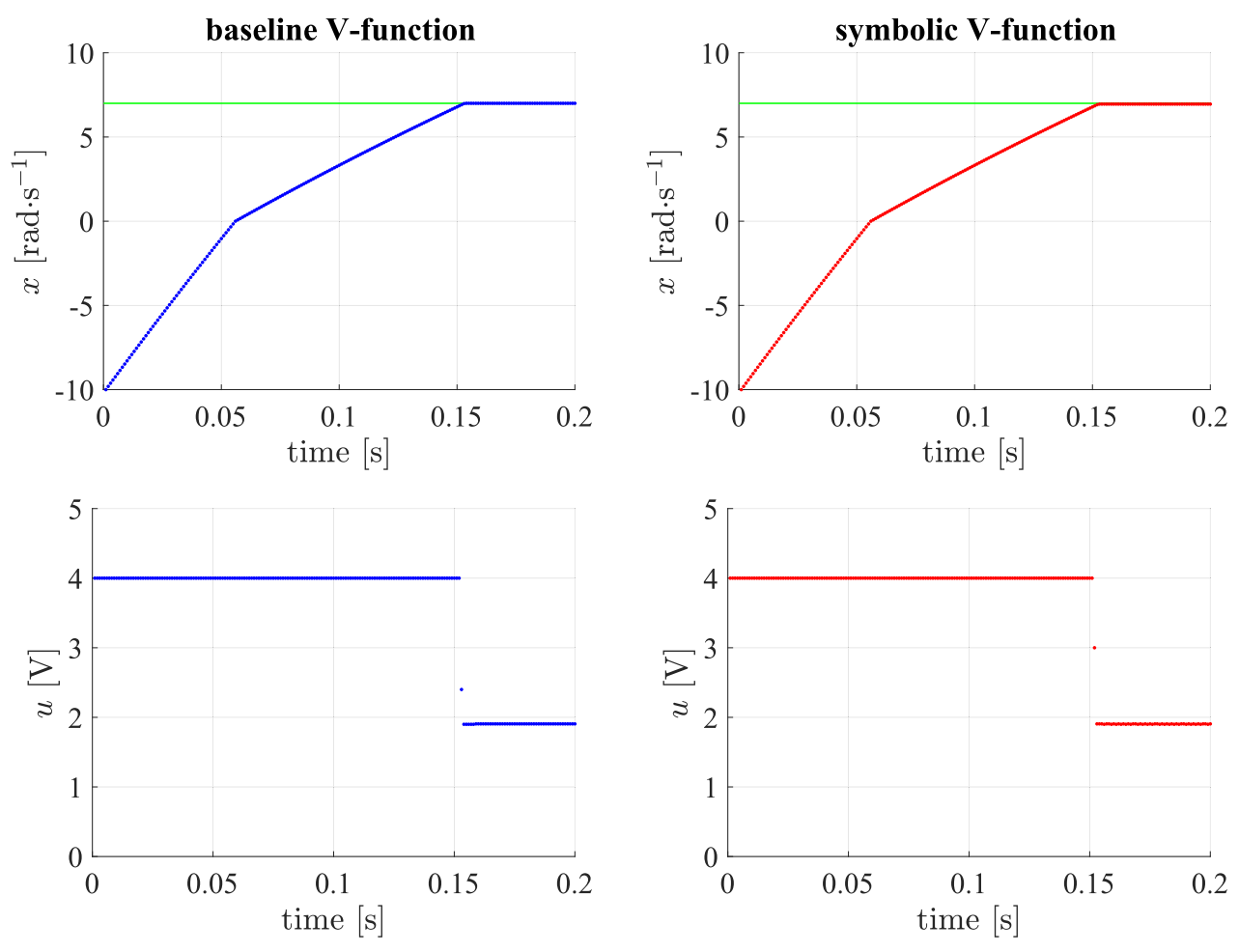

**FIGURE 4. Simulations of the friction compensation problem with the baseline V-function (left) and the symbolic V-function (right) presented in Figure 3b). The upper plots show the state trajectory from $x_0 = -10$ rad·s$^{-1}$. The lower plots show the corresponding control inputs. Only the first 0.2 s of the simulation are shown as the variables remain constant afterward.**

### *B. 1-DOF PENDULUM SWING-UP*

The inverted pendulum (denoted as 1-DOF) consists of a weight of mass $m$ attached to an actuated link that rotates in a vertical plane. The available torque is insufficient to push the pendulum up in a single rotation from many initial states. Instead, from certain states (e.g., when the pendulum is pointing down), it needs to be swung back and forth to gather energy prior to being pushed up and stabilized. The continuous-time model of the pendulum dynamics is:

$$\ddot{\alpha} = \frac{1}{I} \cdot \left[ mgl \sin(\alpha) - b\dot{\alpha} - \frac{K^2}{R}\dot{\alpha} + \frac{K}{R}u \right] \tag{22}$$

where $I = 1.91 \times 10^{-4}$ kg·m$^2$, $m = 0.055$ kg, $g = 9.81$ m·s$^{-2}$, $l = 0.042$ m, $b = 3 \times 10^{-6}$ N·m·s·rad$^{-1}$, $K = 0.0536$ N·m·A$^{-1}$, $R = 9.5\,\Omega$. The angle $\alpha$ varies in the interval $[0, 2\pi]$ rad, with $\alpha = \pi$ rad pointing up, and 'wraps around' so that e.g., a rotation of $5\pi/2$ rad corresponds to $\alpha = \pi/2$ rad. The state vector is $x = [\alpha, \dot{\alpha}]^\top$. The sampling period is $T_s = 0.05$ s, and the discrete-time transitions are obtained by numerically integrating the continuous-time dynamics (22) by using the fourth-order Runge-Kutta method. The control input $u$ is limited to $[-2, 2]$ V, which is insufficient to push the pendulum up in one go.

The control goal is to stabilize the pendulum in the unstable equilibrium defined by the goal state $x_r = [\pi, 0]^\top$, which is expressed by the following reward function:

$$\rho(x, u, f(x, u)) = -|x_r - x|^\top Q \tag{23}$$

with $Q = [0.5, 0]^\top$ a weighting vector to adjust the relative importance of the angle and angular velocity.

Figure 5 and Table 2 present the statistical results obtained from 30 independent runs. Figure 5 shows that the SVI and SPI methods achieve comparable performance, while the direct method fails.

Figure 6 shows an example of a well-performing symbolic V-function found through SR, compared to a baseline V-function calculated using the numerical approximator [13]. The symbolic V-function is smoother than the numerical baseline, as seen from the level curves and the state trajectory. The difference is particularly notable near the goal state, which is a significant advantage of the proposed method.

Figure 7 shows a simulation with the symbolic V-function and an experiment with the real system [5]. The trajectory of the control signal $u$ on the real system shows the typical bang-bang nature of optimal control, which illustrates that SR found a near-optimal value function.

**TABLE 2. Performance of the symbolic methods on the 1-DOF problem. The performance of the baseline V-function is $R_\gamma = -9.346$, BE = 0.0174, $S$ = 100 %.**

| SNGP | direct | SPI | SVI |
|---|---|---|---|
| $R_\gamma$ [–] | −26.083 | −10.187 | −10.013 |
| BE [–] | 0.478 | 0.242 | 0.615 |
| $S$ [%] | 0 | 100 | 100 |
| | [0, 0 (30)] | [81.3, 100 (27)] | [93.8, 100 (28)] |
| **MGGP** | **direct** | **SPI** | **SVI** |
| $R_\gamma$ [–] | −26.083 | −10.487 | −9.917 |
| BE [–] | 0.776 | 0.797 | 0.623 |
| $S$ [%] | 0 | 100 | 100 |
| | [0, 6.25 (2)] | [0, 100 (17)] | [0, 100 (29)] |

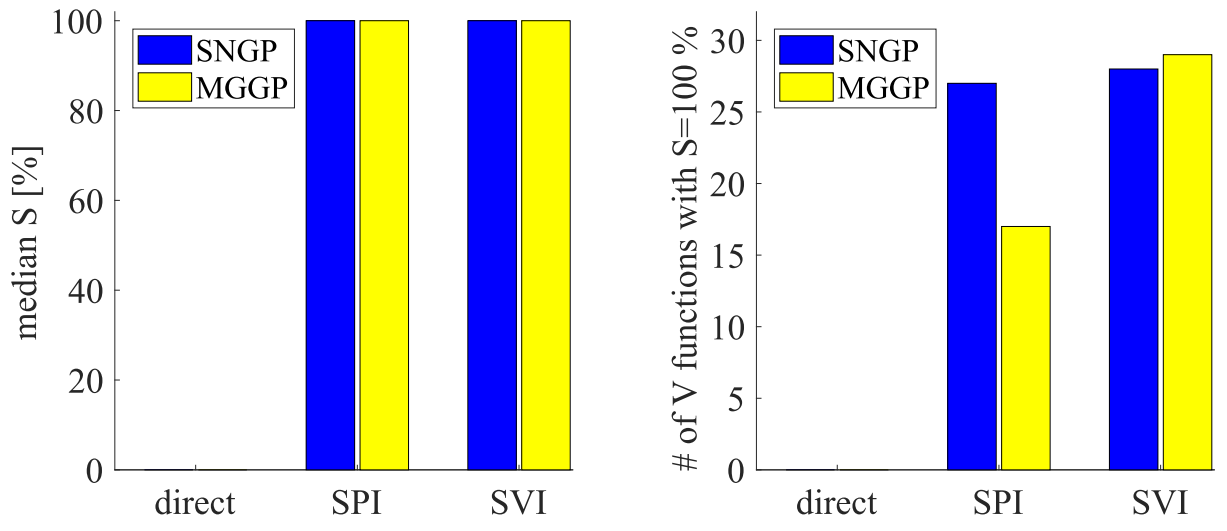

(a) (b)

**FIGURE 5.** Performance on the 1-DOF problem: (a) median $S$, (b) the number of runs in which a V-function with $S = 100\,\%$ was found.

Figure 6 depicts a surface plot of the symbolic V-function constructed by the `SVI-SNGP` method. This function is described by the following analytic expression:

$$\begin{aligned}V(x) = {}& 1.7\times10^{-5}(10x_2-12x_1+47)(4.3\times10^{-2}x_2\\&-3.5x_1+11)^3-7.1\times10^{-4}x_2-4.6x_1-8.2\times10^{-6}\\&(4.3\times10^{-2}x_2-3.5x_1+11)^3(0.2x_1+0.3x_2-0.5)^3\\&-9.8\times10^{-3}(0.4x_1+0.1x_2-1.1)^6+11(0.1x_1-1.5)^3\\&+11((0.6x_1+6.3\times10^{-2}x_2-1.7)^2+1)^{0.5}+8.7\times10^{-6}\\&((10x_2-12x_1+47)^2(4.3\times10^{-2}x_2-3.5x_1+11)^6+1)^{0.5}\\&+0.3((1.1x_1+0.4x_2-3.3)^2+1)^{0.5}+(3.9\times10^{-3}\\&(4.3\times10^{-2}x_2-3.5x_1+11)^2(0.2x_1+0.3x_2-0.5)^2+1)^{0.5}\\&+6.5\times10^{-5}((1.2x_1+14x_2-10)^2(9.1\times10^{-2}x_2-2.9x_1\\&+0.5((9.1\times10^{-2}x_2-2.9x_1+8.3)^2+1)^{0.5}+7.8)^2+1)^{0.5}\\&-5.5\times10^{-2}(4.3\times10^{-2}x_2-3.5x_1+11)(0.2x_1+0.3x_2\\&-0.5)-1.7((3.6x_1+0.4x_2-11)^2+1)^{0.5}-2((x_1-3.1)^2\\&+1)^{0.5}-1.3\times10^{-4}(1.2x_1+14x_2-10)(9.1\times10^{-2}x_2\\&-2.9x_1+0.5((9.1\times10^{-2}x_2-2.9x_1+8.3)^2+1)^{0.5}\\&+7.8)+23. \qquad (24)\end{aligned}$$

This example shows that symbolic V-functions are compact, analytically tractable, and easy to plug into other algorithms. The number of parameters in the example is 100.

### C. 2-DOF SWING-UP

The double pendulum (denoted as 2-DOF) is described by the following continuous-time fourth-order nonlinear model:

$$M(\alpha)\ddot{\alpha} + C(\alpha,\dot{\alpha})\alpha + G(\alpha) = u \qquad (25)$$

with $\alpha = [\alpha_1, \alpha_2]^\top$ the angular positions of the two links, $u = [u_1, u_2]^\top$ the control input, which are the torques of the two motors, $M(\alpha)$ the mass matrix, $C(\alpha,\dot{\alpha})$ the Coriolis and centrifugal forces matrix and $G(\alpha)$ the gravitational forces vector. The state vector contains the angles and angular velocities: $x = [\alpha_1, \dot{\alpha}_1, \alpha_2, \dot{\alpha}_2]^\top$. The angles $\alpha_1$, $\alpha_2$ vary in the interval $[-\pi, \pi)$ rad and wrap around. The angular velocities $\dot{\alpha}_1$, $\dot{\alpha}_2$ are restricted to the interval $[-2\pi, 2\pi)$ rad·s$^{-1}$ using saturation. Matrices $M(\alpha)$, $C(\alpha,\dot{\alpha})$ and $G(\alpha)$ are defined by:

$$M(\alpha) = \begin{bmatrix} P_1 + P_2 + 2P_3\,\cos(\alpha_2) & P_2 + P_3\,\cos(\alpha_2) \\ P_2 + P_3\,\cos(\alpha_2) & P_2 \end{bmatrix},$$

**TABLE 3.** Double pendulum parameters.

| Model parameter | Symbol | Value | Unit |
|---|---|---|---|
| Link lengths | $l_1, l_2$ | 0.4, 0.4 | m |
| Link masses | $m_1, m_2$ | 1.25, 0.8 | kg |
| Link inertias | $I_1, I_2$ | 0.0667, 0.0427 | kg·m$^2$ |
| Center of mass coordinates | $c_1, c_2$ | 0.2, 0.2 | m |
| Damping in the joints | $b_1, b_2$ | 0.08, 0.02 | kg·s$^{-1}$ |
| Gravitational acceleration | $g$ | 9.8 | m·s$^{-2}$ |

**TABLE 4.** Results obtained on the 2-DOF problem. The performance of the baseline V-function is $R_\gamma = -80.884$, $BE = 8 \times 10^{-6}$, $S = 23\,\%$.

| SNGP | `direct` | `SPI` | `SVI` |
|---|---|---|---|
| $R_\gamma$ [–] | −89.243 | −85.607 | −81.817 |
| BE [–] | 4.23 | 2.00 | 5.79 |
| $S$ [%] | 15.4 | 38.5 | 53.8 |
| | [7.7, 23.1 (14)] | [0, 100 (4)] | [7.7, 100 (4)] |
| **MGGP** | `direct` | `SPI` | `SVI` |
| $R_\gamma$ [–] | −84.739 | −84.116 | −82.662 |
| BE [–] | 5.19 | 1.98 | 3.29 |
| $S$ [%] | 23.1 | 26.9 | 69.2 |
| | [0, 30.8 (1)] | [0, 100 (5)] | [7.7, 100 (2)] |

**TABLE 5.** Magnetic manipulation (Magman) system parameters.

| Model parameter | Symbol | Value | Unit |
|---|---|---|---|
| Ball mass | $m$ | $3.200 \times 10^{-2}$ | kg |
| Viscous damping | $b$ | $1.613 \times 10^{-2}$ | N·s·m$^{-1}$ |
| Empirical parameter | $c_1$ | $5.520 \times 10^{-10}$ | N·m$^5$·A$^{-1}$ |
| Empirical parameter | $c_2$ | $1.750 \times 10^{-4}$ | m$^2$ |

$$C(\alpha,\dot{\alpha}) = \begin{bmatrix} b_1 - P_3\dot{\alpha}_2\sin(\alpha_2) & -P_3(\dot{\alpha}_1+\dot{\alpha}_2)\sin(\alpha_2) \\ P_3\dot{\alpha}_1\sin(\alpha 2) & b_2 \end{bmatrix},$$

$$G(\alpha) = \begin{bmatrix} -F_1\sin(\alpha_1) - F_2\sin(\alpha_1+\alpha_2) \\ -F_2\sin(\alpha_1+\alpha_2) \end{bmatrix}$$

with $P_1 = m_1\,c_1^2 + m_2\,l_1^2 + I_1$, $P_2 = m_2\,c_2^2 + I_2$, $P_3 = m_2\,l_1\,c_2$, $F_1 = (m_1\,c_1 + m_2\,l_2)g$ and $F_2 = m_2 c_2\,g$. The meaning and values of the system parameters are given in Table 3. The transition function $f(x, u)$ is obtained by numerically integrating (25) using the fourth-order Runge-Kutta method. The sampling period is $T_s = 0.01$ s.

The control goal is to stabilize the two links in the upper equilibrium, which is expressed by the following quadratic reward function:

$$\rho(x, u, f(x, u)) = -(x_r - x)^\top Q(x_r - x) \qquad (26)$$

with the desired goal state $x_r = [0, 0, 0, 0]^\top$ and $Q = \text{diag}([1, 0, 1.2, 0])$ a weighting matrix to specify the relative importance of the angles and angular velocities.

Figure 8 and Table 4 present the statistical results obtained from 30 independent runs.

### D. MAGNETIC MANIPULATION

Magnetic manipulation has several advantages compared to traditional robotic manipulation approaches. It is contactless, which opens new possibilities for actuation on a micro scale and in environments where it is not possible to use conventional actuators. In addition, magnetic manipulation is not constrained by the robot arm morphology, and it is less constrained by obstacles.

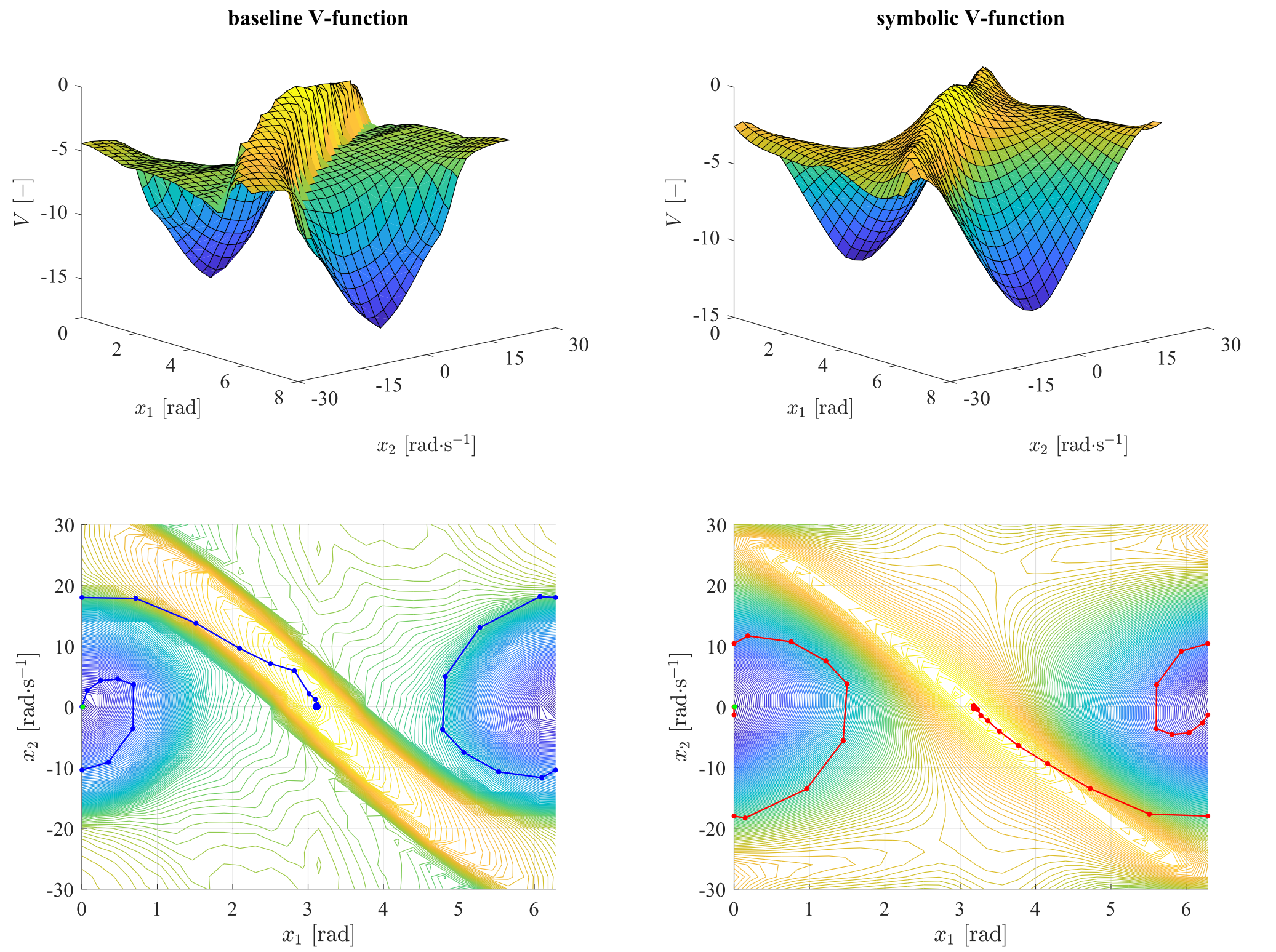

**FIGURE 6.** Baseline and symbolic V-function produced by the SVI−SNGP method on the 1-DOF problem. The symbolic V-function is smoother than the numerical baseline V-function, which can be seen from the level curves and the state trajectory, particularly near the goal state.

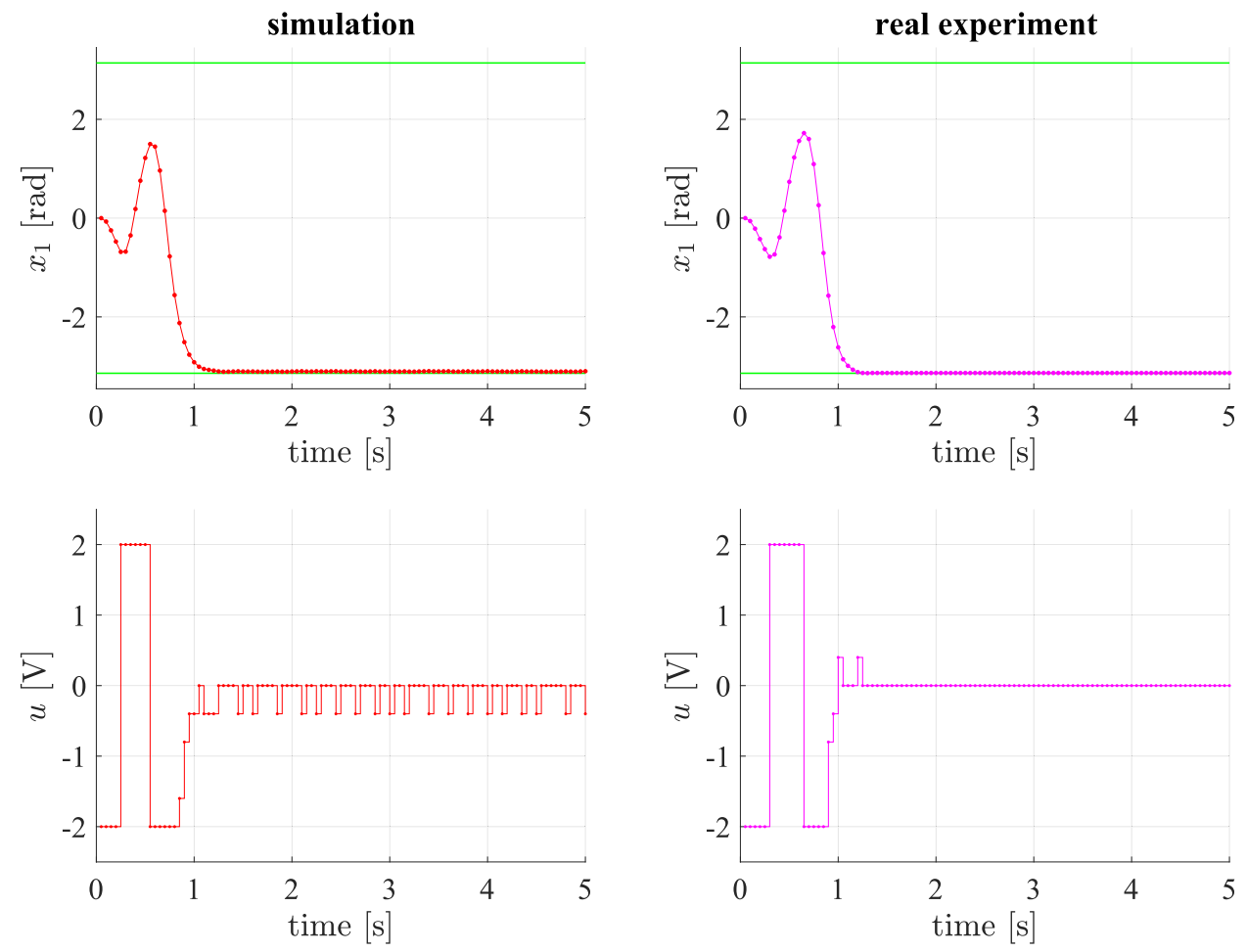

**FIGURE 7.** An example of a well-performing symbolic V-function found with the SVI−SNGP method for the 1-DOF problem, used in a simulation (left) and on the real system (right). The performance of the SVI method is near-optimal both in simulations and in real experiments.

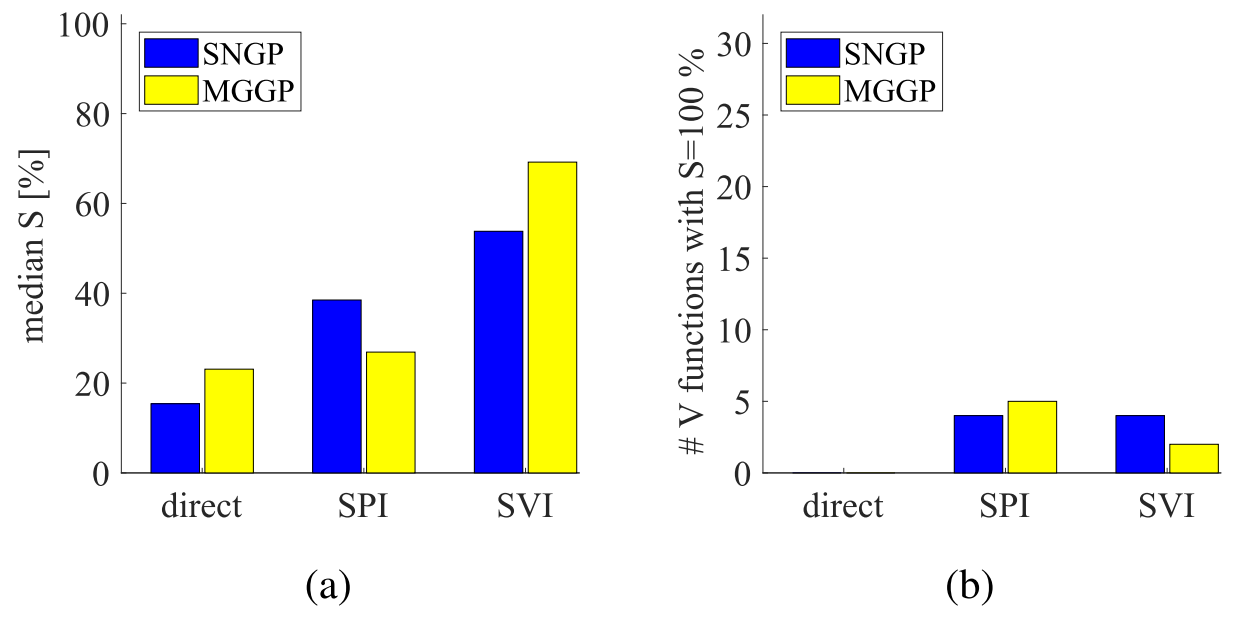

**FIGURE 8.** Performance on the 2-DOF problem: a) median $S$, b) the number of runs, out of 30, in which a V-function achieving $S = 100\,\%$ was found.

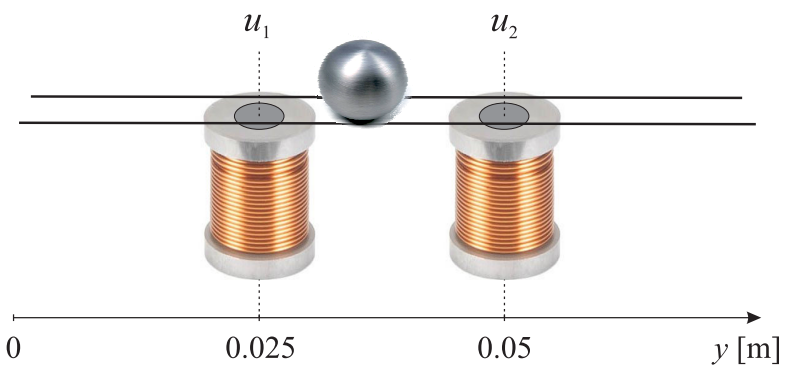

**FIGURE 9.** Magman schematic.

A schematic of a magnetic manipulation setup [35] with two coils is shown in Figure 9. A steel ball freely rolls on a rail mounted above two electromagnets positioned at 0.025 m and 0.05 m. The current through the coils is controlled to dynamically shape the magnetic field above the magnets and so to accurately and quickly position the ball at a desired setpoint.

The horizontal acceleration of the ball is given by:

$$\ddot{y} = -\frac{b}{m}\dot{y} + \frac{1}{m}\sum_{i=1}^{2} g(y, i)\, u_i \tag{27}$$

with

$$g(y, i) = \frac{-c_1\ (y - 0.025i)}{\left((y - 0.025i)^2 + c_2\right)^3}. \tag{28}$$

Here, $y$ denotes the position of the ball, $\dot{y}$ its velocity and $\ddot{y}$ the acceleration. With $u_i$ the current through coil $i$, $g(y, i)$ is the nonlinear magnetic force equation, $m$ the ball mass, and $b$ the viscous friction of the ball on the rail. The model parameters are listed in Table 5.

State $x$ is given by the position and velocity of the ball. The control goal is to stabilize the ball at the goal state $x_r = [0.01, 0]$. The reward function is defined by:

$$\rho(x, u, f(x, u)) = -(x_r - x)^\top Q(x_r - x) \tag{29}$$

with the matrix $Q = \text{diag}([5, 0])$ specifying the relative importance of the ball's position and velocity.

Figure 12 and Table 6 present the statistical results obtained from 30 independent runs. Figure 10 shows an example of a well-performing symbolic V-function found through SR, compared to the baseline V-function with basis functions [13]. The symbolic V-function is smoother and it has only 77 parameters compared to 729 parameters of the basis-functions approximator.

**TABLE 6. Results obtained on the Magman problem. The performance of the baseline V-function is $R_\gamma = -0.0097$, BE = $1.87 \times 10^{-4}$, $S$ = 100 %.**

| SNGP | direct | SPI | SVI |
|---|---|---|---|
| $R_\gamma$ | −9.917 | −0.010 | −0.011 |
| BE | 0.623 | 0.084 | 0.00298 |
| $S$ | 100 | 100 | 100 |
| | [7.14, 100 (27)] | [100, 100 (30)] | [100, 100 (30)] |
| MGGP | direct | SPI | SVI |
| $R_\gamma$ | −0.164 | −0.010 | −0.169 |
| BE | 0.004 | 15.74 | 0.061 |
| $S$ | 14.3 | 100 | 0 |
| | [0, 100 (5)] | [0, 100 (16)] | [0, 100 (4)] |

The state and control action trajectories simulated with the symbolic and baseline V-functions from Figure 10 are presented in Figure 11. The symbolic one performs well; however, the way it approaches the goal state is suboptimal. This result demonstrates the trade-off between the complexity and the smoothness of the V-function.

### E. VALUE ITERATION WITH NEURAL NETWORK APPROXIMATOR

Other types of V-function representations than the symbolic and basis-function ones can be used. Here, we show an analysis of neural network-based approximators when plugged into the V-iteration method. We used the Matlab implementation of the feed-forward neural network, the `fitnet` function.

Neural networks have many (hyper)parameters that must be tuned to solve a particular problem. To provide a reasonably fair analysis, we tested the neural networks with various topologies and settings of two learning control parameters. Four topologies with two hidden layers having 8, 12, 20, and 42 neurons in each of them were tested. These topologies represent models of different complexity with roughly 100, 200, 500, and 2000 parameters to be tuned (we use $W$ to denote the complexity). The other two control parameters are the maximum number of training epochs before the training is stopped, $E \in \{1000, 2000, 5000, 10000\}$, and the maximum number of validation checks before the training stops, $F \in \{20, 50\}$. In each SVI iteration, the training data set is randomly split into training and validation sets, and the neural network training stops when there is no improvement in the validation performance for the last $F$ training epochs. The hidden layers' neurons used the hyperbolic tangent activation function and the output neuron the pure linear activation function. The gradient descent with momentum and adaptive learning rate backpropagation, the `traingdx` method, was used to train the network's weights.

For each control problem, the NN approximators were tested with all 32 combinations of the above hyperparameters. The results are summarized in Table 7. Only the $S$ performance metric, as the most important one, is used to assess the NN approximators' performance. The results for the best performing configuration $[E, F]$ in terms of the number of runs that produced a 100 % correctly working V-function are presented for each NN approximator's complexity $W$.

## V. DISCUSSION

### A. PERFORMANCE OF METHODS

The SPI and SVI methods are able to produce V-functions allowing to successfully solve the underlying control task (indicated by the maximum value of $S$ equal to 100 %) for all the problems tested. They also clearly outperform the direct method. The best performance was observed on the 1-DOF problem (SVI-SNGP and SVI-MGGP generate 28 and 29 V-functions with S = 100 %, respectively) and the Magman (both SPI-SNGP and SVI-SNGP generate 30 V-functions with S = 100 %). However, we observe MGGP is worse than SNGP, particularly when used in SPI on the 1-DOF and Magman problems and in SVI on Magman. This can be attributed to the fact that MGGP does not penalize for a misplacement of the maximum of the V-function model. Note that a wrong position of the V-function's maximum might prevent reaching the goal state.

The performance of all methods was significantly worse on the 2-DOF problem where the SR methods found a V-function that works perfectly in simulations (i.e., S = 100%) in 2 to 5 runs out of 30. The baseline numerical V-function exhibited even worse performance as only 3 out of all simulations started from 13 initial states successfully ended up in the neighborhood of the goal state (i.e., S = 23%). That is a much lower success rate than the median success rate obtained with the SR methods. This can be attributed to the rather sparse coverage of the state space since the approximator was constructed using a regular grid of $11 \times 11 \times 11 \times 11$ triangular basis functions. Note, however, that sampling the state space by using a coarse grid is often necessary for higher-dimensional problems. The number of samples grows exponentially with the state space dimension, leading to prohibitive memory and computational demands for fine grids. The results show that the SR methods are able to generate reliable results even if only sparsely sampled training data are available.

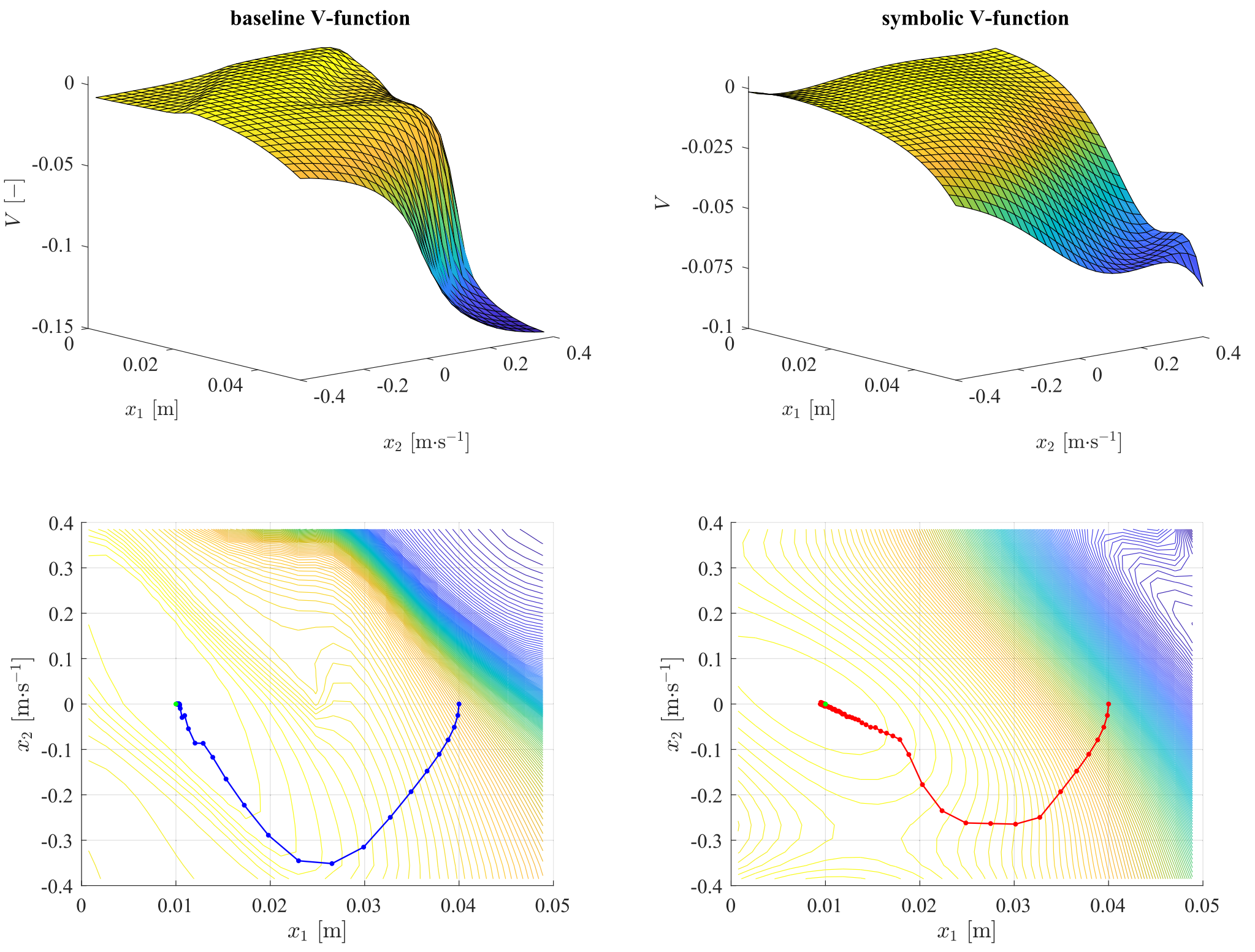

**FIGURE 10.** Baseline and symbolic V-function produced by the `SVI-SNGP` method on the Magman problem. The symbolic V-function is smoother than the numerical baseline V-function, and it performs the control task well. However, the way of approaching the goal state by using the symbolic V-function is inferior to the trajectory generated with the baseline V-function. This example illustrates the trade-off between the complexity of the V-function and the ability of the algorithm to find the intricate details on the V-function surface that matter for the performance.

**TABLE 7.** Performance of neural network-based V-function approximators. The $S$ performance metric is presented for the best neural network configuration $[E, F]$ per approximator's complexity $W$. For the sake of easy comparison, the results obtained with SNGP and MGGP are copied from Table 2, Table 4, and Table 6.

| | | | $S$: 1-DOF | | | | $S$: 2-DOF | | | | $S$: Magman |
|---|---|---|---|---|---|---|---|---|---|---|---|
| SNGP | | | 100, [94, 100 (28)] | SNGP | | | 54, [8, 100 (4)] | SNGP | | | 100, [100, 100 (30)] |
| MGGP | | | 100, [0, 100 (29)] | MGGP | | | 69, [8, 100 (2)] | MGGP | | | 0, [0, 100 (4)] |
| Neural network | | | | Neural network | | | | Neural network | | | |
| $W$ | $E$ | $F$ | | $W$ | $E$ | $F$ | | $W$ | $E$ | $F$ | |
| 100 | 1000 | 50 | 100, [0, 100 (25)] | 100 | 2000 | 50 | 0, [0, 100 (2)] | 100 | 2000 | 50 | 79, [0, 100 (11)] |
| 200 | 1000 | 50 | 100, [63, 100 (27)] | 200 | 10000 | 50 | 0, [0, 100 (3)] | 200 | 5000 | 50 | 79, [0, 100 (8)] |
| 500 | 1000 | 50 | 100, [13, 100 (26)] | 500 | 10000 | 50 | 0, [0, 100 (3)] | 500 | 5000 | 20 | 86, [0, 100 (5)] |
| 2000 | 2000 | 50 | 100, [0, 100 (28)] | 2000 | 10000 | 50 | 0, [0, 100 (3)] | 2000 | 5000 | 50 | 46, [0, 100 (2)] |

Interestingly, the `direct` method implemented with SNGP was able to find several perfect V-functions with respect to $S$ on the Magman. On the contrary, it completely failed to find such a V-function on the 2-DOF and even on the 1-DOF problem. We observed that although the 1-DOF and Magman systems both have a 2D state space, the 1-DOF problem is harder for the symbolic methods in the sense that the V-function has to be very precise in certain regions of the state space to allow for successful closed-loop control. This is not the case in the Magman problem, where V-functions that only roughly approximate the optimal V-function can perform well.

Overall, the two SR methods, SNGP and MGGP, performed well, although SNGP was better on the 1-DOF and Magman problem. Note, however, that a thorough comparison of SR methods was not a primary goal of the experiments. We have also not tuned the control parameters of the algorithms at all, and it is quite likely that if the parameters of the algorithms were optimized, their performance would improve.

**TABLE 8. Experiment parameters.**

| Problem | Friction compensation | 1-DOF | 2-DOF | Magman |
|---|---|---|---|---|
| $n$ | 1 | 2 | 4 | 2 |
| $\mathcal{X}$ | $[-10, 10]$ | $[0, 2\pi] \times [-30, 30]$ | $[-\pi, \pi] \times [-2\pi, 2\pi] \times [-\pi, \pi] \times [-2\pi, 2\pi]$ | $[0, 0.05] \times [-0.4, 0.4]$ |
| $x_r$ | 7 | $[\pi\ 0]$ | $[0\ 0\ 0\ 0]$ | $[0.01\ 0]$ |
| $m$ | 1 | 1 | 2 | 2 |
| $\mathcal{U}$ | $[-4, 4]$ | $[-2, 2]$ | $[-3, 3] \times [-1, 1]$ | $[0, 0.6] \times [0, 0.6]$ |
| $n_u$ | 41 | 11 | 9 | 25 |
| $U$ | $\{-4, -3.8, \ldots, 4\}$ | $\{-2, -1.6, \ldots, 2\}$ | $\{-3, 0, 3\} \times \{-1, 0, 1\}$ | $\{0, 0.15, 0.3, 0.45, 0.6\} \times \{0, 0.15, 0.3, 0.45, 0.6\}$ |
| $n_x$ | 121 | 961 | 14641 | 729 |
| $\gamma$ | 0.95 | 0.95 | 0.95 | 0.95 |
| $n_s$ | 7 | 16 | 13 | 14 |
| $T_s$ [s] | 0.001 | 0.05 | 0.01 | 0.01 |
| $T_{\text{sim}}$ [s] | 1 | 5 | 10 | 3 |
| $\varepsilon$ | 0.05 | $[0.1,\ 1]^\top$ | $[0.1,\ 1,\ 0.1,\ 1]^\top$ | $[0.001,\ 1]^\top$ |
| $T_{\text{end}}$ [s] | 0.01 | 2 | 2 | 1 |

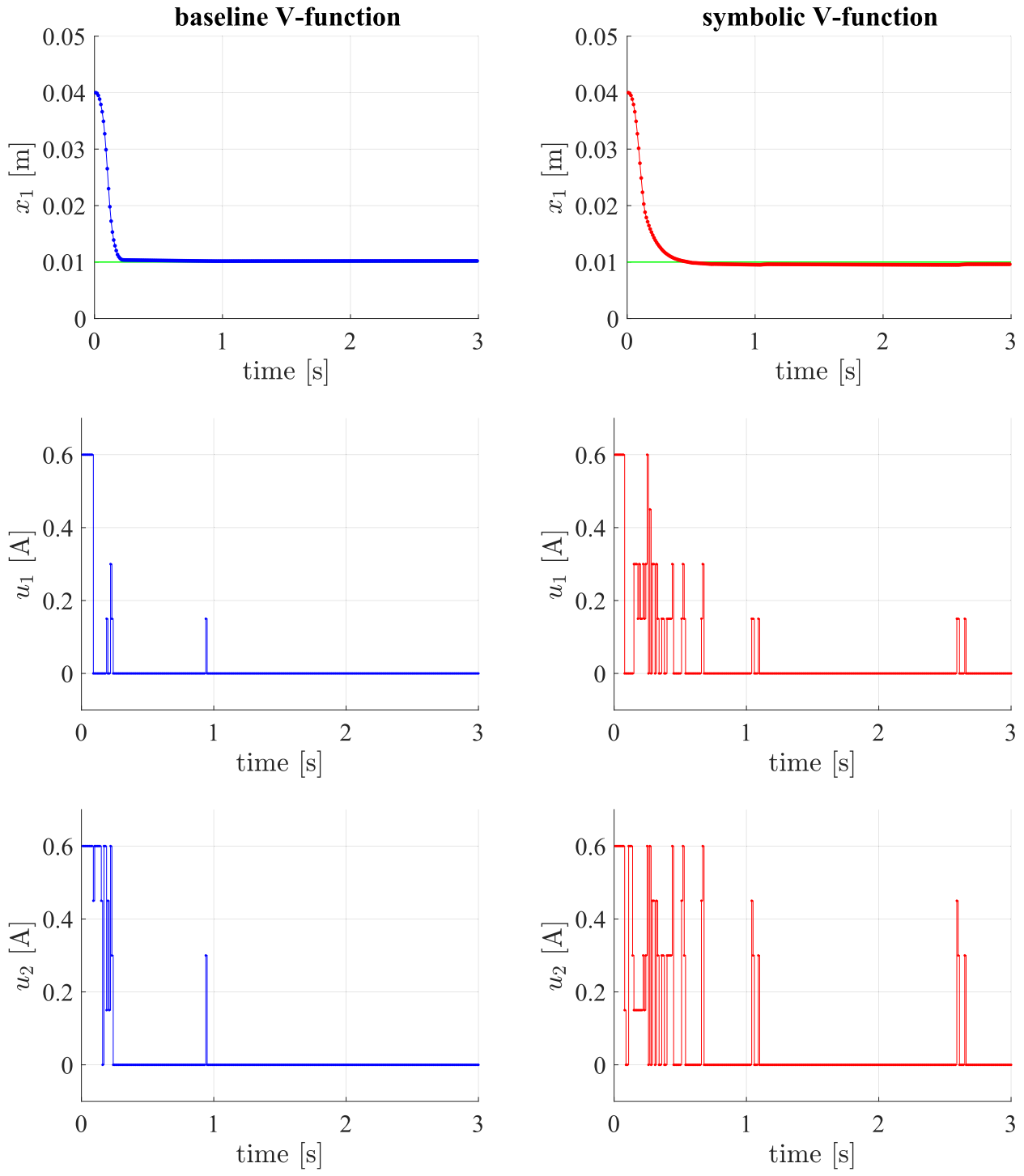

**FIGURE 11. Simulations with the baseline V-function (left) and the symbolic V-function (right) found with the SVI-SNGP method on the Magman problem.**

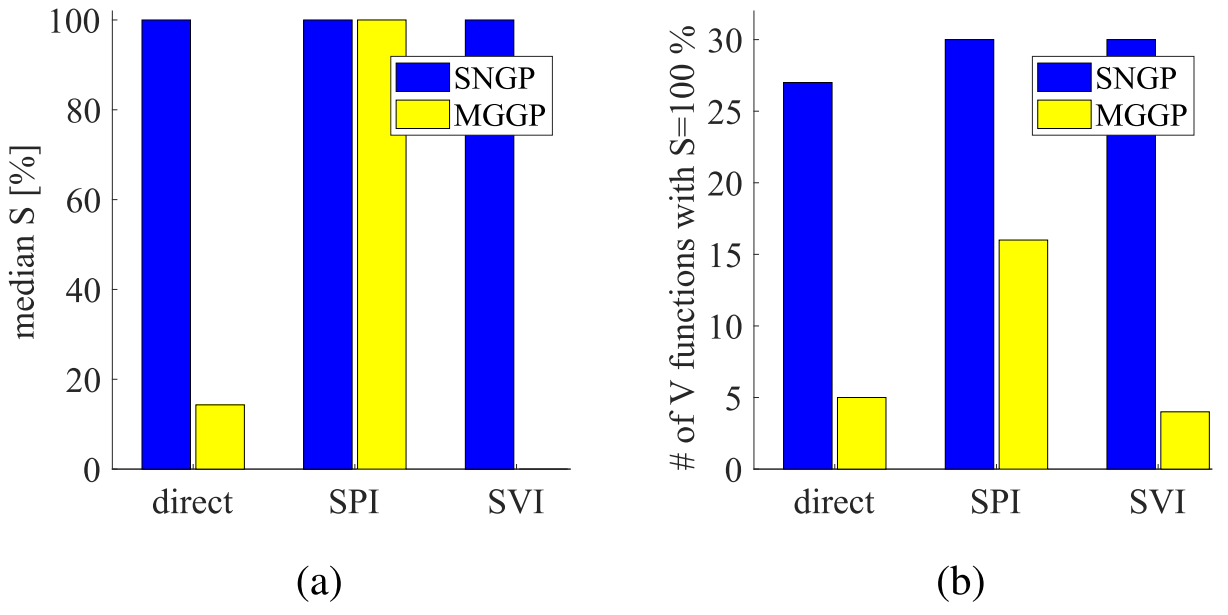

**FIGURE 12. Performance on the Magman problem: a) median $S$, b) the number of runs, out of 30, in which a V-function achieving $S = 100\,\%$ was found.**

## B. COMPARISON WITH NEURAL NETWORK-BASED APPROXIMATORS

NN approximators worked best on the 1-DOF problem, where they achieved performance comparable to the SR methods. Here, larger networks worked slightly better than the smaller ones as they exhibited more stable performance across all possible configurations tested.

On the 2-DOF problem, the most difficult one, the best NN approximator was able to produce a V-function with $S = 100\,\%$ in 3 runs out of 30. Again, larger networks learning for a higher number of epochs performed slightly better than the smaller ones. However, the SR methods are significantly better than the neural networks in the median $S$ value, which is 54 % for SNGP and 69 % for MGGP, compared to 0 % for the neural networks. This means that the SR methods were able to deliver V-functions working correctly for more than 50 % of the $n_s$ initial states in at least 15 runs. NN approximators are much worse in this respect as only up to 10 runs ended up with non-zero (mostly much smaller than 50 %) $S$ metric. Interestingly, a significantly larger network topology with $W = 2000$ did not lead to performance improvement.

On the Magman problem, the overall best neural network performance was observed for $W = 100$. Here, the neural networks worked comparably with MGGP. Both methods were much worse than SNGP in the median $S$ value and the number of runs producing 100 % correct V-function. Interestingly, larger NN topologies only worsened the performance. Like the MGGP, the NN approximator does not penalize models for an incorrectly positioned maximum. In fact, it is not possible to incorporate this kind of desired model's properties into the gradient-based learning algorithm we used.

To conclude, this analysis shows that the SR methods, especially SNGP, outperform the NN approximators. Even more so because the parameters of the NN approximators were tuned for each individual control problem, while the SR methods were run with the same setting on all the problems.

### C. NUMBER OF PARAMETERS

One of the advantages of the proposed symbolic methods is the compactness of the value functions, which can be demonstrated, for instance, on the 1-DOF problem. The symbolic value function found by using the `SVI-SNGP` method (Figure 6, right) has 100 free parameters, while the baseline numerically approximated value function has 961 free parameters.

### D. EASE OF USE

The proposed methods do not require a large amount of expert knowledge in order to be applied to the particular problem at hand. The main parameters of the GP method are the set of elementary functions that are sufficient for creating diverse nonlinear features and the maximum complexity of the models. It is not difficult to choose these parameters. In this work, we used a rather small function set consisting of three simple arithmetical operators $\{*, +, -\}$ and three univariate functions {square, cube, bent identity}. Depending on the problem, the function set can be arbitrarily enriched, for example, by adding trigonometric functions. Furthermore, it is easy to use additional prior knowledge and constraints in SR methods to generate models with desired properties, see [36], [37].

The complexity of the evolved V-functions is defined in terms of the maximum number of features and their maximum depth. We used the maximum feature depth of 7 and the maximum number of features of 10. However, a heuristic guideline for setting up these parameters is that if more complex models are needed, then this could be effectively achieved by increasing the number of features rather than by increasing the maximum feature's depth. There are two reasons for that, (1) the maximum depth of 7 is sufficient to represent complex nonlinear features, and (2) keeping a rather small feature's depth prevents from uncontrolled bloat of evolved expressions that is a severe issue in genetic programming [38].

Although the heuristics mentioned above can guide the setting of the parameters to achieve better results, the methods' performance is not particularly sensitive to the precise choice of these parameters. Neither are the proposed methods dependent on any particular SR algorithm. This was demonstrated while testing the methods with two different GP algorithms that are conceptually very different and have a different set of control parameters. Both GP algorithms were run with rather standard parameter settings inspired by the works using these algorithms for other problems. No parameter tuning was performed in this work. In principle, hyperparameter tuning algorithms can be applied to SR. However, these approaches are extremely computationally expensive.

### E. POST-PROCESSING AND ANALYSIS

Models in the form of analytic expressions are more expressive than, for example, the weights of neural networks and are amenable to further analysis. One can verify properties of the model, such as monotonicity on a specific interval, positions of extremes, symmetry w.r.t. input variables, or other domain-specific properties. Approaches based on satisfiability modulo theory solvers have been used for this kind of formal constraint verification in the literature [37].

Furthermore, the contribution of the individual features to the overall performance of the analytic V-function can be assessed and used to select the most significant features by using, for example, backward elimination. Over-simplified models can be further refined by adding new features to improve the current model's accuracy. This is hard to do with numerical approximators such as neural networks or basis function expansions.

Finally, given a moderate number of internal parameters, it is possible to efficiently fine-tune them using global optimization techniques such as pattern search [39] and evolution strategies [40], [41].

### F. COMPUTATIONAL COMPLEXITY

The time needed for a single run of the `SVI`, `SPI` or `direct` method ranges from several minutes for the illustrative example to around 24 hours for the 2-DOF problem on a standard desktop PC. The running time of the algorithm increases linearly with the size of the training data. However, the size of the training data set may grow exponentially with the state space dimension. In this article, we have generated the data on a regular grid. Other data generation methods are part of our future research. For high-dimensional problems, SR has the potential to be computationally more efficient than numerical approximation methods such as deep neural networks.

## VI. CONCLUSION

We have proposed three methods based on SR to construct an analytic approximation of the V-function in a Markov decision process. The methods were experimentally evaluated on four nonlinear control problems: one first-order system, two second-order systems, and one fourth-order system.

The main advantage of the proposed approach is that it produces smooth, compact V-functions, even if only sparsely sampled training data are available. The models produced are mathematically tractable and easy to analyze. The number of their parameters is an order of magnitude smaller than in a basis-function approximator. The control performance in simulations and experiments on a real setup is excellent. Moreover, the approximator in the form of a set of analytic expressions allows for postprocessing and fine-tuning. It can also be easily reused within and plugged into other algorithms. For example, smooth policy derivation methods exploit the analytic nature of the symbolic V-function model [12].

No parameter tuning was used for the SR methods. We consider it as an essential advantage of SR methods that they work well without any particular tuning. This contrasts with, e.g., deep neural network methods that often require substantial tuning before one gets them even to converge in a given RL problem. The most significant current limitation of the approach is its high computational complexity.

In our future work, we will evaluate the method on higher-dimensional problems, where we expect a considerable benefit over numerical approximators in terms of computational complexity. To this end, we will investigate smart methods for generating the training data. We will also examine the use of input–output models instead of state-space models and closed-loop stability analysis methods for symbolic value functions. We will also develop techniques to incrementally control the complexity of the symbolic value function depending on its performance.

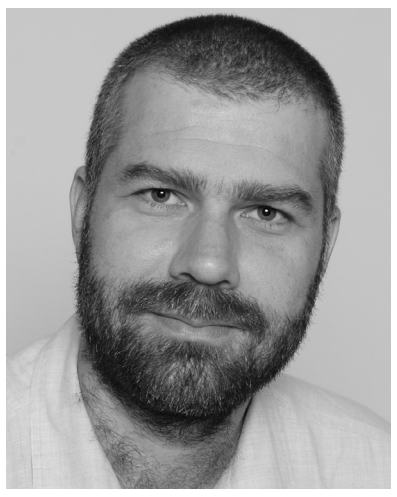

**JIŘÍ KUBALÍK** received the M.Sc. degree in computer science and the Ph.D. degree in artificial intelligence and biocybernetics from Czech Technical University (CTU) in Prague, in 1994 and 2001, respectively. He is currently a Researcher with the Czech Institute of Informatics, Robotics and Cybernetics, CTU in Prague. He is a (co)author of more than 30 articles in this area. His research has mainly been focused on various types of evolutionary computation techniques and their applications to hard optimization problems.

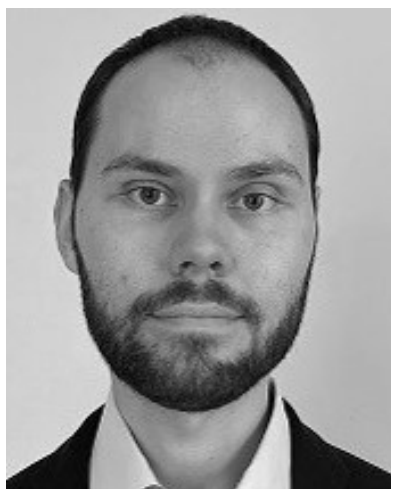

**ERIK DERNER** (Student Member, IEEE) received the M.Sc. degree (Hons.) in artificial intelligence and computer vision from Czech Technical University (CTU) in Prague, where he is currently pursuing the Ph.D. degree. His research interests include sample-efficient methods for mobile robotics, genetic programming, reinforcement learning, and computer vision. He currently focuses on long-term autonomy in robot control and navigation. The central topic in his research is the use of SR to automatically construct nonlinear models of dynamic systems.

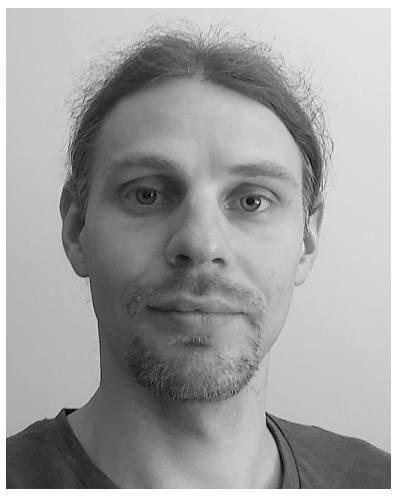

**JAN ŽEGKLITZ** received the M.Sc. degree in artificial intelligence from Czech Technical University (CTU) in Prague, Czech Republic, where he is currently pursuing the Ph.D. degree with the Department of Cybernetics, Faculty of Electrical Engineering. His research interests include the field of evolutionary algorithms and genetic programming. The main research focus is on multi-gene genetic programming.

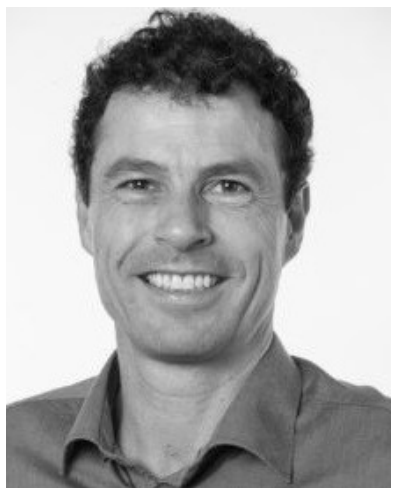

**ROBERT BABUŠKA** (Member, IEEE) received the M.Sc. degree (Hons.) in control engineering from Czech Technical University in Prague, in 1990, and the Ph.D. degree *(cum laude)* from the Delft University of Technology, The Netherlands, in 1997. He had faculty appointments with Czech Technical University in Prague and the Electrical Engineering Faculty, TU Delft. He is currently a Full Professor of intelligent control and robotics at the Department of Cognitive Robotics, Faculty 3mE, TU Delft. In the past, he made seminal contributions to the field of nonlinear control and identification with the use of fuzzy modeling techniques. His current research interests include reinforcement learning, adaptive and learning robot control, nonlinear system identification, and state estimation. He has been involved in the applications of these techniques in various fields, ranging from process control to robotics and aerospace.